\documentclass[12pt,longbibliography,eprint]{revtex4-1}
\usepackage{graphicx}
\usepackage{epstopdf}
\usepackage{hyperref}

\begin{document}
% $Id: RNimages.tex,v 1.105 2015/12/15 12:49:00 mal Exp $
\preprint{V.M.}
\title{Radon--Nikodym approximation in application
  to image reconstruction.
}
\author{Vladislav Gennadievich \surname{Malyshkin}}
\email{malyshki@ton.ioffe.ru}
\affiliation{Ioffe Institute, Politekhnicheskaya 26, St Petersburg, 194021, Russia}

\date{November 3, 2015}

\begin{abstract}
\begin{verbatim}
$Id: RNimages.tex,v 1.105 2015/12/15 12:49:00 mal Exp $
\end{verbatim}
For an image  pixel information
can be converted to the moments
of some basis $Q_k$, e.g. Fourier--Mellin,  Zernike,
monomials, etc.
Given sufficient number of
moments pixel information can be completely recovered,
for insufficient number of moments only partial
information can be recovered
and the image reconstruction is, at best, of interpolatory type.
Standard approach is to present interpolated value
as a linear combination of basis functions,
what is equivalent to least squares expansion.
However, recent progress in numerical stability of moments estimation
allows  image information to be recovered from moments in
a completely different manner, applying Radon--Nikodym type of expansion,
what gives the result as a ratio of two quadratic forms of basis
functions.
In contrast with least squares the Radon--Nikodym approach
has oscillation near the boundaries very much suppressed and does not diverge
outside of basis support.
While least squares theory operate with vectors $<fQ_k>$,
Radon--Nikodym theory operates with matrices $<fQ_jQ_k>$,
what make the approach much more suitable to image transforms and
statistical property estimation.
\end{abstract}

\maketitle
%\begin{IEEEkeywords}
%Radon Nikodym, Nevai Operator
%\end{IEEEkeywords}

\section{\label{intro}Introduction}
Image information representation is a fundamental question
of image processing and analysis. Most common basis
is pixel basis. However given basis functions $Q_k$
(e.g. Fourier,  Zernike, orthogonal polynomials\cite{totik}, etc.) 
image pixel information
can be transformed to the moments of the basis\cite{mukundan1998moment,pinoli2014mathematical,honarvar2014image}.
Given sufficient number of moments a complete
one-to-one mapping between pixel and moments information can be established.
However, given limited number of moments a question arise:
how image information can be recovered from moments available.
Most common approach -- representation of the result
in a form of linear combination of basis functions,
what is equivalent to least squares approximation.
However, there is exist a different approach based
on Radon--Nikodym derivatives\cite{kolmogorovFA}
and its special case Nevai Operator\cite{nevai1979orthogonal,nevai},
where the result is represented as a ratio
of two quadratic forms of basis functions.
In contrast with least squares, which operate
on vector moments  $<f Q_k>$ of observable value $f$,
the Radon--Nikodym approach operate with matrices $<f Q_k Q_j>$.
Given recent progress in numerical stability of high
order moments calculation\cite{2015arXiv151005510G}
the matrices $<f Q_k Q_j>$ can now be calculated
without any difficulty to a very high order and Radon--Nikodym
become practically applicable to image processing.
This matrix approach, has a number of unique features,
such as suppression of typical for least squares oscillations
near the boundary and improved numerical stability.
In addition to that 
the transition from vector
to matrix allows many image 
transforms to be easily expressed 
in terms of matrix $<f Q_k Q_j>$ transform
and the approach allows to leverage matrix algebra
in application to image processing.

\section{\label{expansion}Basis expansion}
Consider some feature $f$ (e.g. grayscale intensity),
a basis $Q_k(x)$ (in 2D the basis would be $Q_{k_x}(x)Q_{k_y}(y)$)
and the measure $\mu$ (in this paper the measure would be just the sum over the
pixels). The moments are defined as:
\begin{eqnarray}
  <fQ_k>&=&\int\limits_{\mathrm{supp}(\mu)} f Q_k d\mu
  \label{innerprod}
\end{eqnarray}
  The Gramm matrix is defined by the basis and the measure:
\begin{eqnarray}
  G_{ij}=<Q_iQ_j> \label{gramm}
\end{eqnarray}
Then minimization of
mean square difference between $f$ and its approximation $A_{LS}(x)$
obtain standard least squares result:
\begin{equation}
  A_{LS}(x)=\sum_{i,j} Q_i(x)\left(G^{-1}\right)_{ij}<fQ_j> \label{leastsq}
\end{equation}
Radon--Nikodym approximation $A_{RN}(x)$ can be obtained considering
localized at $x_0$ states $\psi_{x_0}(x)$
\begin{eqnarray}
  \psi_{x_0}(x)&=&\frac{\sum_{i,j} Q_i(x)G^{-1}_{ij}Q_j(x_0)}
  {\sqrt{\sum_{i,j}Q_i(x_0)G^{-1}_{ij}Q_j(x_0)}}
  \label{psix0norm}
\end{eqnarray}
and a form of Radon--Nikodym approximation,
Nevai Operator\cite{nevai1979orthogonal,nevai},
then becomes:
\begin{equation}
  A_{RN}(x)=\frac{\sum_{i,j,k,m}Q_i(x) G^{-1}_{ij} <fQ_jQ_k> G^{-1}_{km} Q_m(x)}
  {\sum_{i,j}Q_i(x) G^{-1}_{ij}Q_j(x)}
  \label{RNsimple}
\end{equation}
The main idea is to consider localized at $x_0$ states $\psi_{x_0}(x)$,
which is related to delta-function expanded in $Q_k(x)$ basis
with measure (\ref{innerprod}),
and perform $f$ reconstruction as $f(x_0)\approx \int dx f(x)\psi^2_{x_0}(x)/\int dx \psi^2_{x_0}(x)$. Important, that integration weight $\psi^2_{x_0}(x)$
is always positive what supress oscillations typical for least squares,
where the weight change sign.
The $\psi_{x_0}(x)$ from (\ref{psix0norm})
give exactly Nevai operator (\ref{RNsimple}).
For details and other $\psi_{x_0}(x)$ forms applicable for
Radon--Nikodym estimation see Ref. \cite{2015arXiv151005510G}.
The (\ref{RNsimple}), while is very different from least squares in concept,
uses, nevertheless, almost the same input:
Gramm matrix inverse $G^{-1}$ and $<Q_iQ_kf>$ matrix obtained from $<Q_jf>$ moments. The (\ref{RNsimple}) is a ratio of two polynomial functions.
It was shown in Ref. \cite{malha} that in multi--dimensional signal processing
stable estimators can be only of two quadratic forms ratio and
the (\ref{RNsimple}) is exactly of this form.

Let us apply least squares and Radon--Nikodym expressions
to some real life cases.

\subsection{\label{D1}1D Example: Runge Function.}
Before we start considering 2D images, let us start with
simple 1D example, take 
Runge function
\begin{equation}
  f(x)=\frac{1}{1+25x^2} \label{rungeF}
\end{equation}
Using the measure
\begin{eqnarray}
  <f Q_{k}>&=& \int_{-1}^{1} f(x) Q_k(x) dx
  \label{mom1D}
\end{eqnarray}
and the basis, for numerical stability of calculations,
is chosen as Legendre polynomials $Q_k(x)=P_k(x)$
(given the measure all polynomial bases provide identical results,
but numerical stability of calculations is  different).

The $A_{RN}(x)$ calculation algorithm is this:
given $N$ elements in basis using (\ref{mom1D}) definition
calculate vector moments $<fQ_{m}>$ and $<Q_{m}>$
for $m=[0\dots 2N-1]$.
Then, applying polynomials multiplication operation,
for $j,k=[0\dots N-1]$
obtain matrix moments $<fQ_{j}Q_{k}>$ and $G=<Q_{j}Q_{k}>$
to be used in Eq. (\ref{leastsq}) and (\ref{RNsimple}).

% set output "q.eps" ; set terminal postscript eps size 12cm,8cm enhanced color ; set xrange [-1:1]; set yrange [-0.1:1] ; set xtics 0.2 ; set ytics 0.1 ; set ylabel "f" ;  set xlabel "x" ; l=693; s=0.5;cx=0.7; oo=0.7;ox=0.5; plot   "/tmp/r.txt" using ($1):($2) with lines title "f", "/tmp/r.txt" using ($1):($3) with lines title "Af_{LS}"  lc 3,  "/tmp/r.txt" using ($1):($4) with lines title "Af_{RN}"  lc 2
% set output "q.eps" ; set terminal postscript eps size 12cm,8cm enhanced color ; set xrange [-1:1]; set yrange [-3.5:3.5] ; set xtics 0.2 ; set ytics 0.5 ; set ylabel "df/dx" ;  set xlabel "x" ; l=693; s=0.5;cx=0.7; oo=0.7;ox=0.5; plot   "/tmp/r.txt" using ($1):($5) with lines title "df/dx", "/tmp/r.txt" using ($1):($6) with lines title "ADf_{LS}"  lc 3,  "/tmp/r.txt" using ($1):($7) with lines title "ADf_{RN}"  lc 2 , "/tmp/r.txt" using ($1):($8) with lines title "DAf_{LS}"  lc 5,  "/tmp/r.txt" using ($1):($9) with lines title "DAf_{RN}"  lc 7

\begin{figure}
\includegraphics[width=9cm]{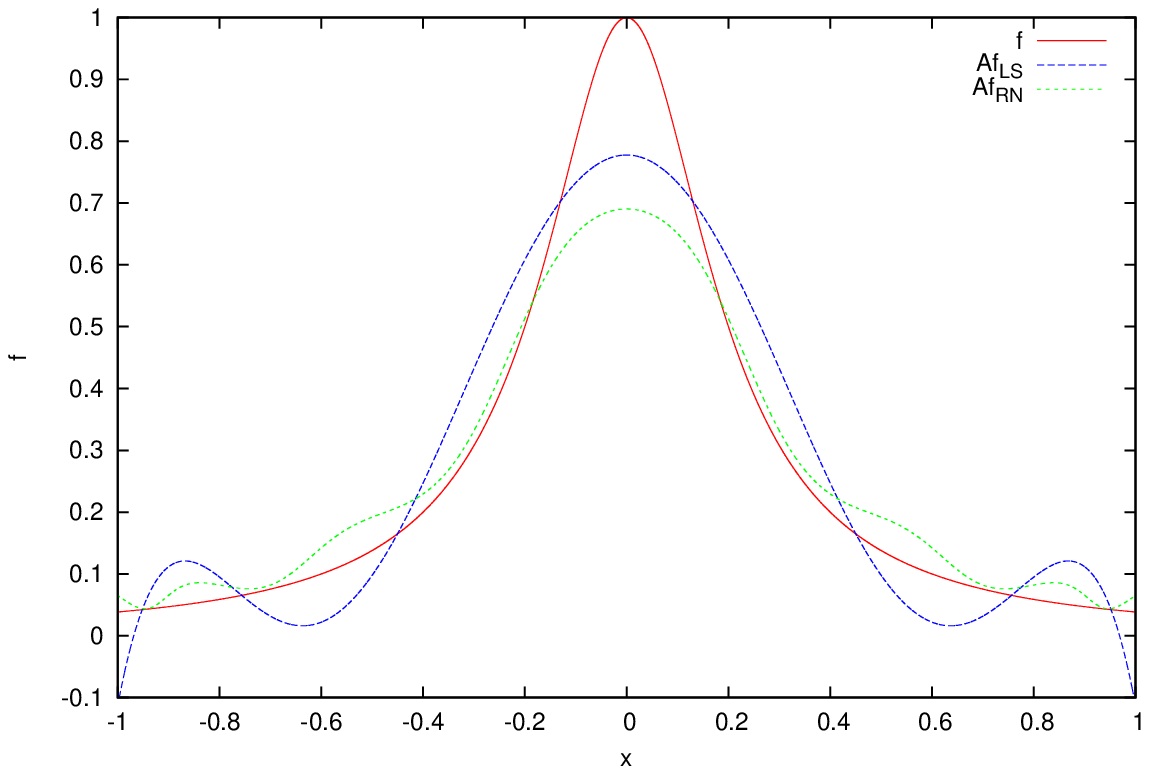}
\includegraphics[width=9cm]{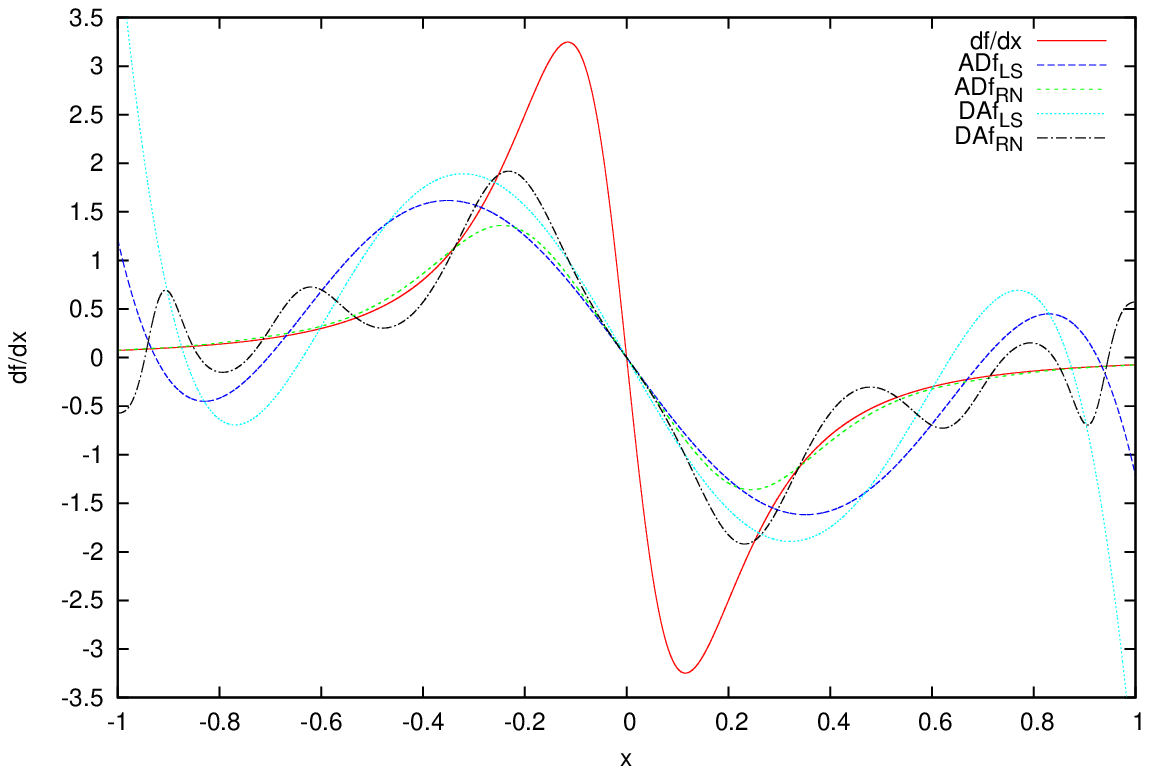}
\caption{\label{fig:fig_runge}
  Top chart: Runge function $f$, least squares approximation $Af_{LS}$
  and Radon--Nikodym approximation $Af_{RN}$.
  Bottom chart:
  Runge function derivative $\frac{df}{dx}$,
  least squares approximation $ADf_{LS}$
  and Radon--Nikodym approximation $ADf_{RN}$ of the derivative $\frac{df}{dx}$.
  Differentiated interpolations of $f$
  for least squares($DAf_{LS}$) and Radon--Nikodym ($DAf_{RN}$),
  that should not be used in applications are also presented as an example.
}  
\end{figure}

In top chart of Fig. \ref{fig:fig_runge}  least squares and Radon--Nikodym
interpolations are presented for $N=7$ and the measure (\ref{mom1D}).
One can see that near edges oscillations are much
less severe, when  Radon--Nikodym approximation as polynomials ratio is used
for the interpolation of $f$.
The major behavior differences
for least square and Radon--Nikodym approximations are:
Least squares have diverging oscillations near
measure support boundaries and tend to infinity with the distance to
measure support increase. Radon--Nikodym have converging
oscillations near
measure support boundaries and tend to a constant with the distance to
measure support increase.

Another, worth to mention point, is related to derivatives calculation.
For this the moments $\left<Q_k \frac{df}{dx}\right>$
should be calculated first,
and only then applying Radon--Nikodym approximation like (\ref{RNsimple}) using derivative moments.
If one, instead of using the $\left<Q_k \frac{df}{dx}\right>$ moments,
would  differentiate $f$ approximation expression (\ref{RNsimple}) directly --
the result will be incorrect. To illustrate the point
in bottom chart of Fig. \ref{fig:fig_runge}  least squares and Radon--Nikodym
interpolations of Runge function derivative
are presented for $N=7$ and measure (\ref{mom1D}).
The differentiated approximations of Runge function
 are also presented.

The code calculating this 1D example is available\cite{polynomialcode},
see the ExampleRungeFunction.scala file.

\subsection{\label{ex2D} 2D example: Lena image.}

Let us consider 2D case of image grayscale intensity interpolation.
Because the example is illustrative,
let us take a sum over image pixels as the measure.
As a basis, in principle, monomials $x^{k_x}y^{k_y}$ can be used,
but for numerical stability reasons the basis should be chosen
as orthogonal functions with respect to some measure.
For an image of $d_x$ on $d_y$
pixels the moments of pixel--dependent grayscale intensity $f(t_x,t_y)$
(here $t_x=[0\dots d_x-1]$ and $t_y=[0\dots d_y-1]$ index pixel number)  are:
\begin{eqnarray}
 && <f Q_{k_x,k_y}>=   \label{momsample} \\
&&  \sum_{t_x=0}^{d_x-1}\sum_{t_y=0}^{d_y-1} f(t_x,t_y)
  Q_{k_x}(t_x/(d_x-1))Q_{k_y}(t_y/(d_y-1))  \nonumber
\end{eqnarray}
In the Eq. (\ref{momsample})
the basis can be  chosen as Legendre or Chebyshev polynomials
shifted to $[0\dots 1]$ interval:
$Q_k(x)=P_k(2x-1)$ and the argument of $Q$ is pixel coordinate
converted to this interval: $x=t_x/(d_x-1)$ and $y=t_y/(d_y-1)$.
When $d_x\to\infty$ $d_y\to\infty$ the Gramm matrix
$<Q_{i_x,i_y} Q_{j_x,j_y}>$ is diagonal because of Legendre polynomials
orthogonality and trivially invertable,
but we used sample--calculated matrix,
because the $d_x$ and $d_y$ can be rather small or 
when the  basis is chosen as Chebyshev polynomials
$Q_k(x)=T_k(2x-1)$.
Note, that for a given measure all polynomial bases
(e.g. Legendre, Chebyshev, monomials) give identical results,
but numerical stability of calculations is drastically different,
because Gramm matrix condition number depend strongly
on basis choice\cite{beckermann1996numerical}. For 
successful application in image reconstrcution
Chebyshev moments see Ref. \cite{mukundan2001image}
and for Legendre moments see Ref. \cite{chiang2014image}.

The numerical library we developed, see\cite{2015arXiv151005510G} Appendix A,
is able to manipulate polynomials 
in Chebyshev, Legendre, Laguerre and Hermite bases directly,
what allows a stable basis to be used
and calculate the moments to a very high order.
Numerical calculations with polynomials in general basis were
introduced in \cite{Barnett} and
similar technique was used in \cite{laurie1979computation} for Gauss quadratures
calculation in Chebyshev basis. In this paper we used general polynomial basis
approach (to achieve numerical stability)
and applied it to Radon--Nikodym approximation calculation.

The $A_{RN}$ calculation algorithm is this:
given image size $d_x$ and $d_y$ and basis dimension $N_x$ and $N_y$
using (\ref{momsample}) definition
calculate vector moments $<fQ_{m_x,m_y}>$ and $<Q_{m_x,m_y}>$
for $m_x=[0\dots 2N_x-1]$ and $m_y=[0\dots 2N_y-1]$.
Then, applying polynomials multiplication operation,
for $j_x,k_x=[0\dots N_x-1]$ and $j_y,k_y=[0\dots N_y-1]$
obtain
matrix moments $<fQ_{j_x,j_y}Q_{k_x,k_y}>$ and $G=<Q_{j_x,j_y}Q_{k_x,k_y}>$
to be used in Eq. (\ref{leastsq}) and (\ref{RNsimple}).

\begin{figure}
\includegraphics[width=7.5cm]{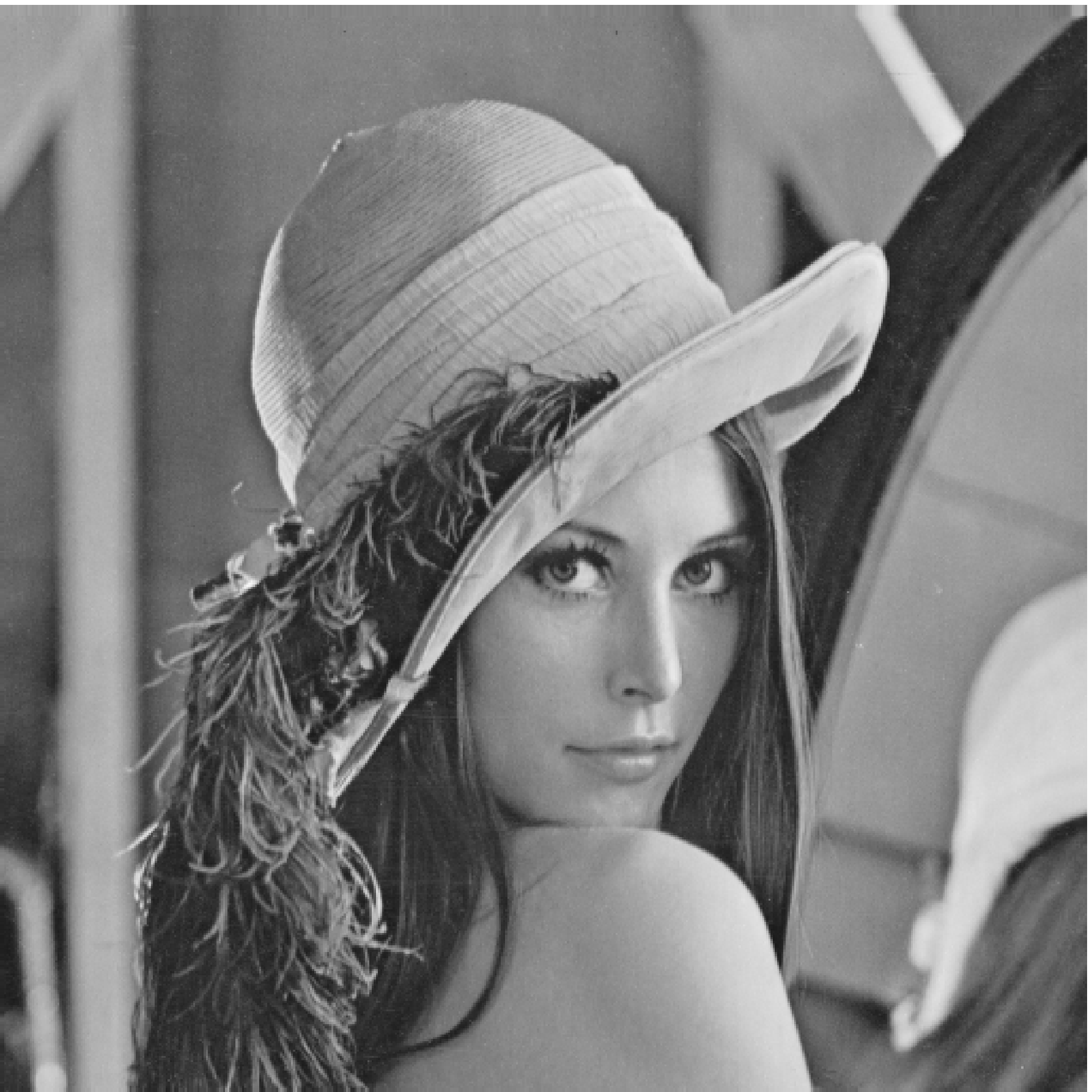}
\includegraphics[width=7.5cm]{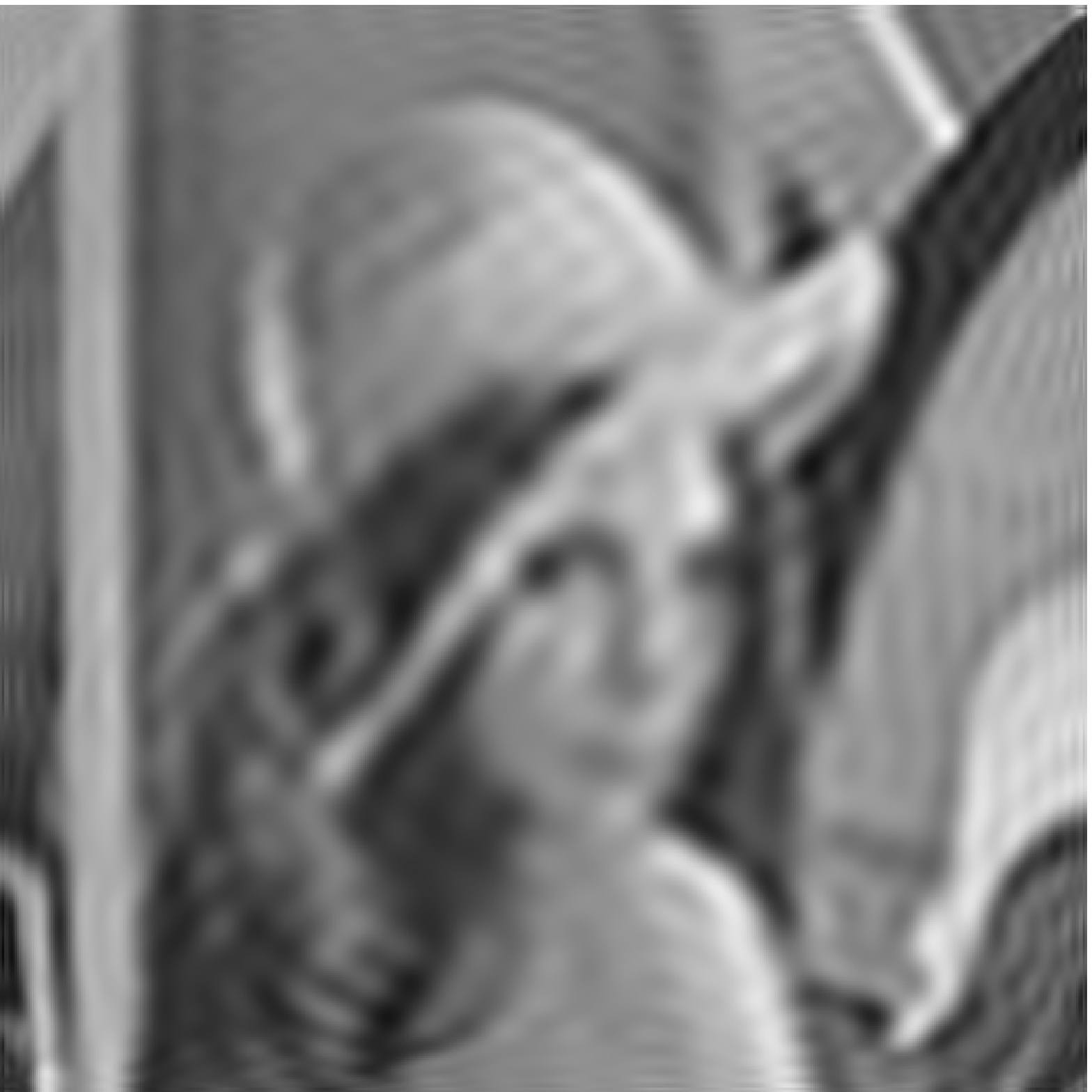}
\includegraphics[width=7.5cm]{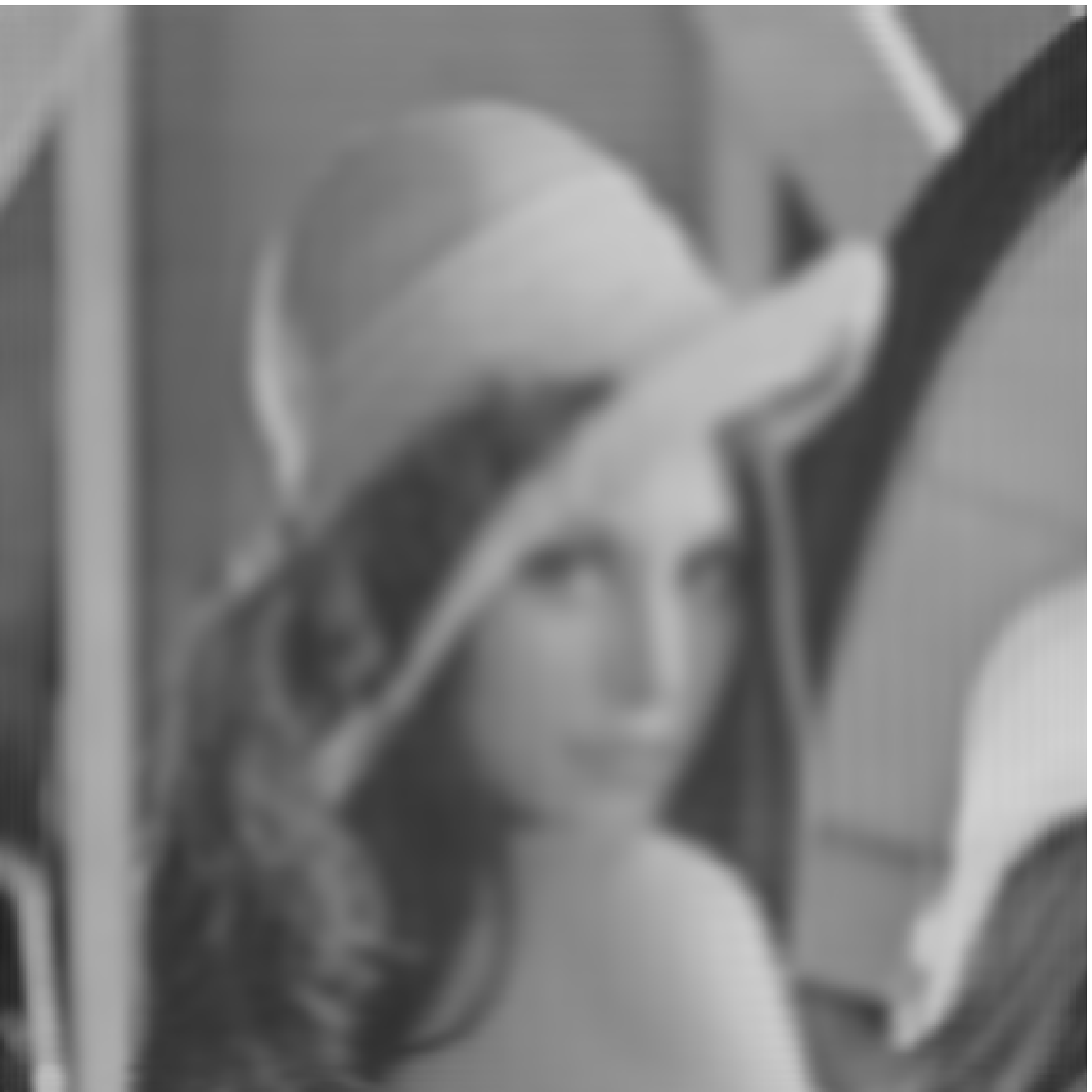}
\caption{\label{fig:lena50}
  Original Lena image(top left), and for $N_x=N_y=50$ least squares(top right) and Radon-Nikodym (bottom). PNG originals are availabe from \cite{2015arXiv151101887G}.
}  
\end{figure}

\begin{figure}
\includegraphics[width=7.5cm]{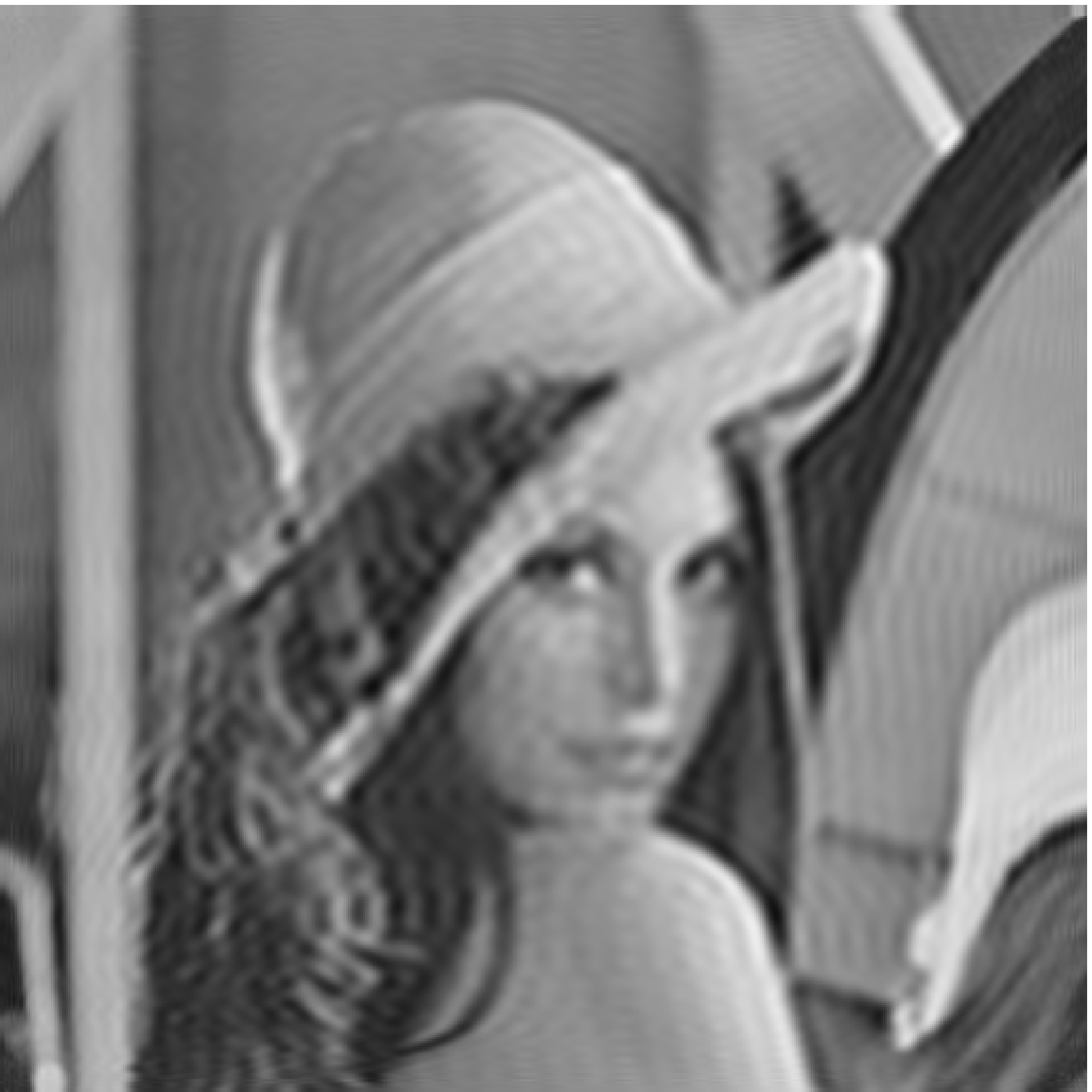}
\includegraphics[width=7.5cm]{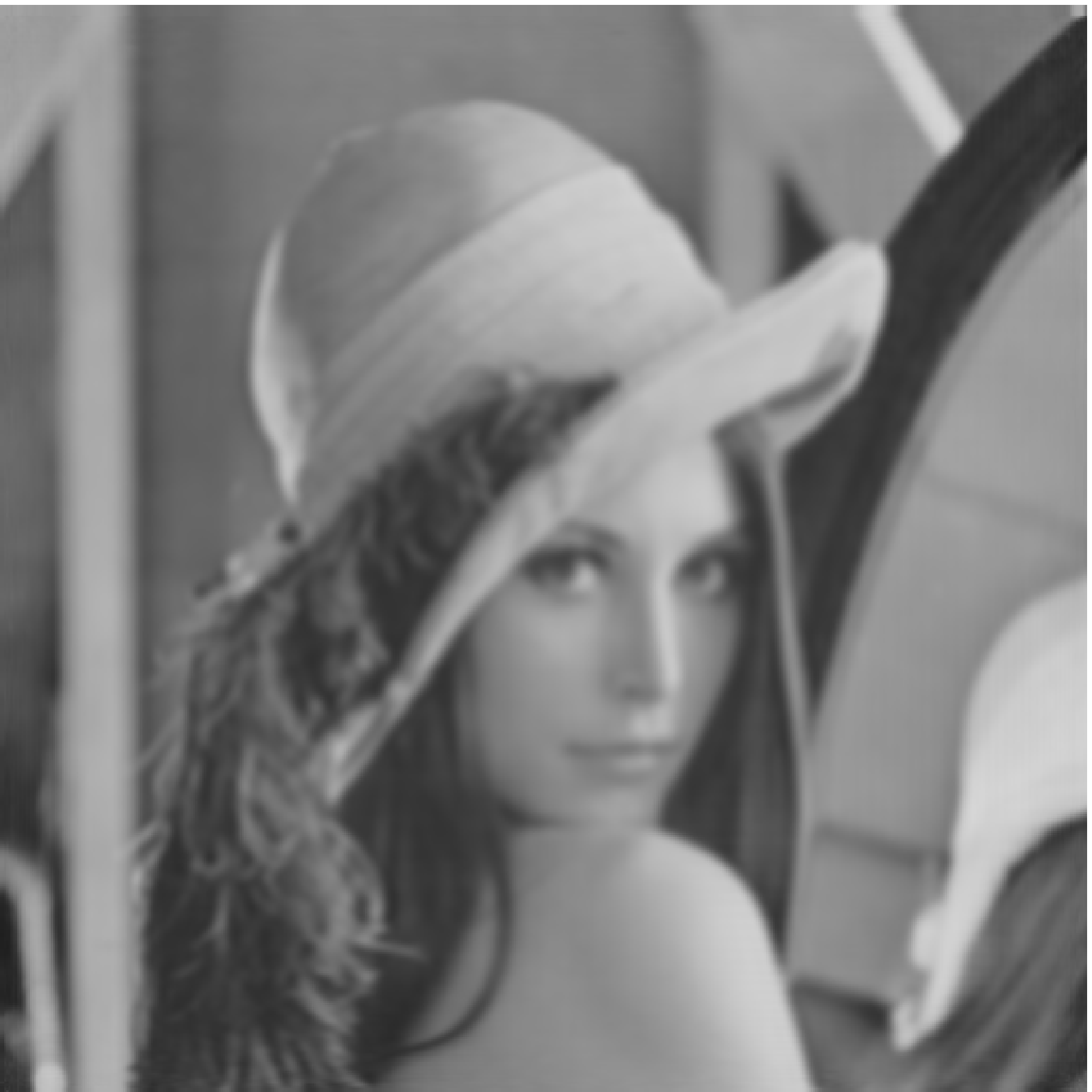}
\caption{\label{fig:lena100}
  Interplation of Lena image for $N_x=N_y=100$. Least squares(left) and Radon-Nikodym (right). PNG originals are availabe from \cite{2015arXiv151101887G}.
}  
\end{figure}

In Fig. \ref{fig:lena50}
we present original 512x512 ($d_x=d_y=512$) grayscale Lena image,
then for $N_x=N_y=50$ apply
least squares (\ref{leastsq})
and Radon--Nikodym   (\ref{RNsimple}) transforms.
Same calculations, but for $N_x=N_y=100$ are presented
in Fig. \ref{fig:lena100}. (The calculations
for $N_x=N_y=100$ are rather slow, because we did not use
any optimization, but the point of the paper is
to demonstrate practical applicability of
Radon--Nikodym type of interpolation and stability of high order
moments calculation when a stable basis is chosen.)

The least squares interpolation, same as in 1D case,
present typical for least squares intensity oscillations near image edges,
while  Radon--Nikodym has these oscillations very much supressed.
Another important feature of Radon--Nikodym is that
it preserves the sign of interpolated function, i.e.
the grayscale intencity $f$ never become negative,
what may happen easily for least squares.
The code calculating this example is available\cite{polynomialcode},
see the ExampleImageInterpolation.scala file.
The calculations have been performed in both: Legendre and Chebyshev bases.
For $N_x=N_y=50$ the results are identical, when interpolated grayscale is converted back to 1-byte values. For $N_x=N_y=100$ the results in two bases are almost identical (indistinguishable visually),
but testing show that in Legendre basis
numerical instabilities just started to show up in multiplication operation,
because of factorial--like coefficients in $P_n(x)P_m(x)$ expansion.
One can expect more instability in Legendre basis at $N_x=N_y>100$
(note, that for given $N$ we calculate $0..2N-1$ moments).
In this sense Chebyshev multiplication
$T_n(x)T_m(x)=\frac{1}{2}T_{n+m}(x)+\frac{1}{2}T_{|n-m|}(x)$ is special because the coeficients of product expansion
do not grow or vanish for large $n ; m$, so Chebyshev products
can be stably calculated to a very high order
and in the same time for discrete measures the Gramm matrix (\ref{gramm}) posess
a good condition number\cite{beckermann1996numerical} in this basis.

\section{\label{spur}Discussion and Natural Basis}

In this paper we present a novel approach to image restoration
from moments: the result is of Radon--Nikodym type where
the result is a ratio of two quadracic forms of basis functions,
and, in case of polynomial bases, is just two polynomials ratio.
This approach,
is based on matrices, not on vectors, what make
calculations significantly more stable.
In a way how Radon--Nikodym approach improved interpolation
of a function, the transition from a vector $<Q_k f>$ 
to matrix $<Q_jQ_k f>$ can similary improve calculations of image
properties, expressible through averages.
Define an average $\overline{f}$ as:
\begin{eqnarray}
  \overline{f}&=&\frac{\mathrm{Spur}\left(\sum_{k}G_{jk}^{-1}<Q_kfQ_l>\right)}{\dim G} \nonumber \\&=&\frac{\sum_{j,k}G_{jk}^{-1}<Q_kfQ_j>}{\dim G}\label{rnspuraver}
\end{eqnarray}
where $\mathrm{Spur}$ is matrix trace (sum of diagonal elements) operator.
The (\ref{rnspuraver}) definition
can be also applied to estimation of an average of products, i.e.
\begin{equation}
  \overline{fg}=\frac{\sum_{j,k,m,i}G_{jk}^{-1}<Q_kfQ_m>G_{mi}^{-1}<Q_igQ_j>}{\dim G}\label{spuraver}
\end{equation}
What allows image features cross--correlation
to be expressed as matrix Spur.
An important feature of the approach is that
many image transforms can be easily expressed
as a transform of matrix $<Q_kfQ_m>$ , what makes
proposed matrix approach extremely practical,
when a stable basis is chosen. Numerical library
providing four stable bases (Legendre Chebyshev, Laguerre, Hermite)
is available from author\cite{polynomialcode}.

And in conclusion we want to mention that
generalized eigenvalues problem
\begin{equation}
  \sum_m<Q_kfQ_m> \psi^{(s)}_m = \lambda^{(s)} \sum_m<Q_kQ_m>\psi^{(s)}_m 
  \label{genev}
\end{equation}
\begin{equation}
  \psi^{(s)}(x)=\sum_m \psi^{(s)}_m Q_m(x)
  \label{psiev}
\end{equation}
when solved\footnote{
  In general case generalized eigenvector problem,
  when scalar product is defined not by a unit matrix, but
  by some other positively defined matrix, Gramm matrix in our case,
  is not any more problematic to solve numerically,
  than regular eigenvalues problem.
  It can be solved using standard, e.g. LAPACK\cite{lapack} routines dsygv,  dsygvd and similar.}
provide a ``natural basis'' of eigenvectors $\psi^{(s)}$
in which both matrices $<Q_kfQ_m>$ and $<Q_kQ_m>$ are simultaneously diagonal.
Besides providing exceptional numerical stability this basis
is a ``natural basis'' for the image,
and can be extremely convenient to store and process image information.
For example, because
\begin{eqnarray}
  <\psi^{(r)} \psi^{(s)}>&=&\delta_{rs}=G_{rs} \\
   <\psi^{(r)} f \psi^{(s)}>&=&\lambda^{(s)}\delta_{rs}
\end{eqnarray}
the Gramm matrix is diagonal in natrural basis ---
the cross-correlation of image features (\ref{spuraver}),
calculated as matrix Spur,
take exceptionally simple form.
This ``natural basis'' can be considered as Radon--Nikodym derivatives
generalization. While Radon--Nikodym derivatives are based on localized at $x_0$
states $\psi_{x_0}(x)$ from (\ref{psix0norm})
the eigenfunctions $\psi^{(s)}$ from (\ref{psiev})
have no such localization constrain
and their localization depend only on image properties.
The value of this ``generalized Radon--Nikodym derivative''
at $\psi^{(s)}$ state is the eigenvalue $\lambda^{(s)}=<\psi^{(s)} f \psi^{(s)}>/<\psi^{(s)} \psi^{(s)}>$. The difference
between Radon--Nikodym and ``generalized Radon--Nikodym''
is similar to conceptual difference\cite{kolmogorovFA}
between Riemann integral,
where the terms are grouped by their closeness in argument--space,
like $\psi_{x_0}(x)$ from (\ref{psix0norm}),
and Lebesgue integral
where the terms are grouped by their closeness in value--space,
like $\psi^{(s)}$ from (\ref{psiev}).
A Lebesgue--type integration using ``generalized Radon--Nikodym''
would look, schematically, like this: For the $f$, defining
Lebesgue integration, solve the (\ref{genev}) problem.
Split interval of $f$ values to a number of $[f_i ; f_i+df]$ intervals.
Then define Lebesgue measure $d\mu$: for every such interval
count the number of eigenvalues $\lambda^{(s)}$
that fall within interval range $f_i\le\lambda^{(s)}< f_i+df$,
this number would be the Lebesgue measure $\mu_i$.
Then Lebesgue integral of some function $\int g(f) d\mu$ is just
$\sum_i g(f_i) \mu_i$. The concept is very similar to the ``density of states''
concept from quantum mechanics, where the density of states is
a number of Hamiltonian eigenvalues that fall within given energy interval.
In practice the Lebesgue--type integration is most often performed
in pixel basis, where the number of pixels with $f$
falling within interval range $f_i\le f< f_i+df$
is considered to be the Lebesgue measure $\mu_i$.
When Lebesgue--type integration
is performed in ``natural basis''
the number of eigenvalues, instead of the number of pixels,
is considered to be the Lebesgue measure $\mu_i$.

\bibliography{LD}

\end{document}